\documentclass[11pt,a4paper]{article}
\usepackage[hyperref]{acl2020}
\usepackage{times}
\usepackage{latexsym}

\usepackage{microtype}
\usepackage{amsmath}
\usepackage{amsfonts}
\usepackage{mathrsfs}
\usepackage{cleveref}
\usepackage[switch]{lineno}
\usepackage{xcolor}
\usepackage{graphicx}
\usepackage{multirow}

\aclfinalcopy

\title{Jointly Optimizing State Operation Prediction and Value Generation for Dialogue State Tracking}

  \author{Yan Zeng \\
  DIRO, Université de Montréal \\
  \texttt{yan.zeng@umontreal.ca} \\\And
  Jian-Yun Nie \\
  DIRO, Université de Montréal \\
  \texttt{nie@iro.umontreal.ca} \\}

\date{}

\begin{document}
\maketitle

\begin{abstract}
We investigate the problem of multi-domain Dialogue State Tracking (DST) with open vocabulary. Existing approaches exploit BERT encoder and copy-based RNN decoder, where the encoder predicts the state operation, and the decoder generates new slot values. 
However, in such a stacked encoder-decoder structure, the operation prediction objective only affects the BERT encoder and the value generation objective mainly affects the RNN decoder. 
In this paper, we propose a purely Transformer-based framework, where a single BERT works as both the encoder and the decoder. In so doing, the operation prediction objective and the value generation objective can jointly optimize this BERT for DST. 
At the decoding step, we re-use the hidden states of the encoder in the self-attention mechanism of the corresponding decoder layers to construct a flat encoder-decoder architecture for effective parameter updating. 
Experimental results show that our approach substantially outperforms the existing state-of-the-art framework, and it also achieves very competitive performance to the best ontology-based approaches.
\end{abstract}

\section{Introduction}

\begin{figure}[t]
\centering
\includegraphics[height=1.7in]{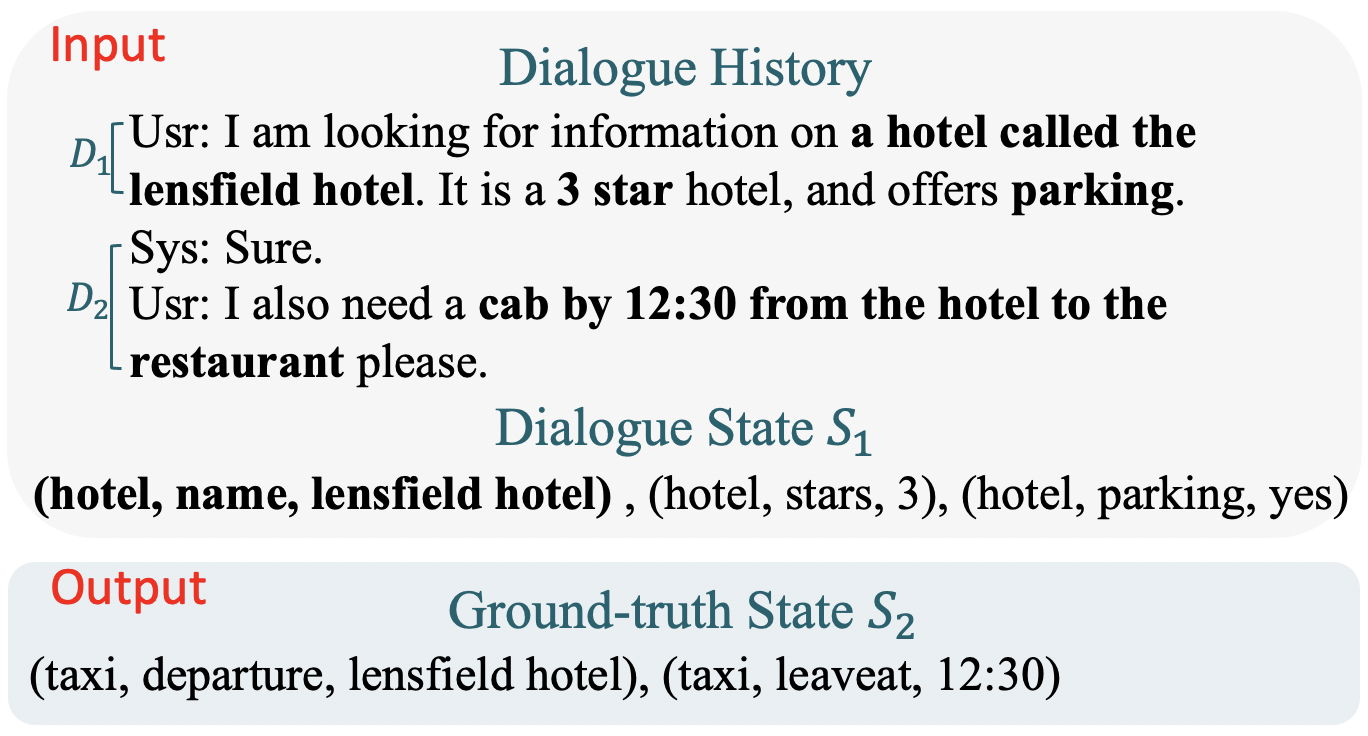}
\caption{An example of multi-domain DST.  
}
\label{Fig:FB_intro}
\end{figure}

Dialogue state tracking (DST) is a core component in task-oriented dialogue systems. Accurate DST is crucial for appropriate dialogue management, where the user intention is an important factor that determines the next system action. Figure \ref{Fig:FB_intro} shows an example of multi-domain DST \cite{budzianowski2018multiwoz}, where the goal is to predict the output state, i.e. (\textit{domain}, \textit{slot}, \textit{value}) tuples given the \textit{dialogue history} and previous \textit{dialog state}.

Studies on DST have started with ontology-based DST \cite{henderson2014word, mrkvsic2017neural}. However, it is often difficult to obtain a large ontology in a real scenario \cite{xu2018end}. Thus, recent studies focus on the open-vocabulary setting \cite{gao2019dialog, zhang2019find, wu2019transferable}, where the possible \textit{values} are not pre-defined and need to be directly extracted / generated from the input. 
The existing state-of-the-art generative framework \cite{kim2019efficient, zhu2020efficient, zeng2020multidomain} decomposes DST into two sub-tasks: State Operation Prediction (SOP) and Value Generation (VG). 
First, the encoder reads the dialogue history and previous dialogue state, and decides whether the value of a (domain, slot) pair needs to be updated. Then, the decoder generates a slot value for the (domain, slot) pair. This approach employs BERT \cite{devlin2018bert} as the encoder and stacks an RNN-based decoder upon BERT outputs. By exploiting BERT as the encoder, this approach has substantially outperformed previous RNN-only framework (RNN encoder and decoder) \cite{wu2019transferable}, and achieved new state-of-the-art performance.

However, the two sub-tasks of DST, i.e. state operation prediction and value generation, are not jointly optimized in this framework, which may lead to a sub-optimal model. Specifically, the SOP objective only affects the BERT encoder, while the VG objective mainly affects the RNN decoder since the stacked encoder-decoder structure makes it less effective in updating the BERT encoder parameters \cite{liu2018generating}. Besides, in the framework, the encoder is pre-trained while the decoder is not. 
This second problem has been observed in \citet{kim2019efficient}, and the proposed solution is to employ two different optimizers for the encoder and the decoder in the training process. The solution further separates the encoder and the decoder, or the SOP and VG process, making it even more difficult to make a global optimization.

Therefore, we propose a purely Transformer-based framework for DST that exploits a single BERT as both the encoder and the decoder. When used as encoder, it processes state operation prediction as in previous works. When using it as decoder, we utilize different input representations to denote the target (decoding) side and left-to-right self-attention mask (i.e. attention is allowed only to previous positions) to avoid information leak. Therefore, the SOP objective and the VG objective affect both the encoder and the decoder, i.e. jointly fine-tuning this BERT for DST. Furthermore, instead of a stacked encoder-decoder structure as in previous studies, whose parameters cannot be effectively updated, we propose 
a flat structure by re-using the hidden states of the encoder in the self-attention mechanism of the corresponding decoder layers to make parameter updating in encoder more effective.

When directly employing the above purely Transformer-based generative framework, we observe, however, that the model performance drops sharply. The possible reason is that DST is not a genuine generation task that usually needs to cope with the entire input. For example, generating a dialogue response needs to be consistent with the dialogue history, or translating a sentence usually needs to translate each word on the encoder (input) side. In contrast, in DST, the value to generate only relates to a very small fraction of the model input (within dialogue history and previous dialogue state) that usually consists of one or few tokens. In this case, asking the decoder to take into account all the encoder outputs may blur the focus. 

To solve the problem, we make the following adaptation by borrowing ideas from the existing state-of-the-art framework of DST: For a specific (domain, slot) pair, our decoder only re-uses the hidden states of the most relevant inputs. After an exhaustive search, our experiments show that re-using dialogue of the current turn and the slot state for the specific (domain, slot) pair yields the best performance, which substantially outperforms the previous framework and only needs a half of the training iterations. 

The contributions of this work are as follows \footnote{Codes are available at \url{https://github.com/zengyan-97/Transformer-DST/}.}:
\begin{itemize}
\item We propose a purely Transformer-based generative framework for DST. The framework jointly optimizes the state operation prediction and value generation process. It also has a flat encoder-decoder architecture allowing for more effective parameter updating.

\item Our method (Transformer-DST) achieves state-of-the-art performance, with joint goal accuracy of 54.64\% and 55.35\% on MultiWOZ 2.0 and MultiWOZ 2.1, improving the existing generative framework by 2.9\% and 2.3\% respectively. 

\item Our model can converge to its best performance much faster and in a more stable manner than the existing framework. This shows the efficiency and robustness of the joint optimization of operation prediction and value generation.

\end{itemize}

\section{Related Work}
\label{sec:related}
Traditional DST approaches rely on ontology. They assume that the possible values for each slot are pre-defined in an ontology and the problem of DST can be simplified into a value classification/ranking task for each slot  \cite{henderson2014word, mrkvsic2017neural, zhong2018global, ren2018towards, ramadan2018large}. These studies showed the great impact of ontology on DST. A recent work \cite{shan2020contextual} combining ontology and contextual hierarchical attention has achieved high performance on MultiWOZ 2.1. In real application situations, however, one cannot always assume that an ontology is available \cite{xu2018end, wu2019transferable}. In many cases, slot values are discovered through the conversation rather than predefined (e.g. taxi departure time). 

Open-vocabulary DST addresses this problem: it tries to extract or generate  a slot value from the dialogue history \cite{lei2018sequicity, gao2019dialog, ren2019scalable}. In this work, we focus on open-vocabulary DST. However, many of the existing approaches did not efficiently perform DST since they generate a value for each slot at every dialogue turn \cite{wu2019transferable}. In contrast, some works \cite{ren2019scalable, kim2019efficient} used a more efficient approach that decomposes DST into two successive sub-tasks: state operation prediction and value generation. Many recent works \cite{zhu2020efficient, zeng2020multidomain} are built upon this approach. However, these models do not jointly optimize the two sub-tasks, which may lead to a sub-optimal model 
since the performance of one process directly influences the performance of another process. 
For example, only when a slot indeed needs updating would generating a new value for it be meaningful. We will solve this problem in this work.   

Studies on open-vocabulary DST have started with RNN-based encoder-decoder architecture. For example, \citet{wu2019transferable} encodes the dialogue history using a bi-directional GRU and decodes the value using a copy-based GRU decoder. Some recent studies have used pre-trained BERT as the encoder \cite{zhang2019find, ren2019scalable, kim2019efficient} to leverage
the rich general linguistic features encoded in BERT. The existing state-of-art generative framework, SOM-DST\cite{kim2019efficient}, utilizes BERT as the encoder to predict state operations, and an RNN decoder stacked upon BERT to generate values. This gives rise to the separate training problems of the two processes we discussed earlier. Different from this approach, we use a single BERT as both the encoder and the decoder, and jointly optimize it with the SOP and VG objective. Furthermore, we also propose a flat encoder-decoder structure instead of stacked encoder-decoder, by re-using hidden states of the encoder in the self-attention mechanism of the corresponding decoder layers. This will lead to more effective updating of the
encoder parameters \cite{liu2018generating}.  

Recently, Tripy \cite{heck2020trippy}, an extractive approach with 2 memory networks, and SimpleTOD \cite{hosseini2020simple}, a language model, have also achieved high performance on MultiWoz 2.1. SimpleTOD adapts the idea of  ``text-to-text'' \cite{radford2019language, raffel2019exploring, khashabi2020unifiedqa} specifically to task-oriented dialogue, and applies multi-task learning on 3 subtasks of task-oriented dialogue including DST. DST can thus benefit from the two other sub-tasks. In addition, a much larger pre-trained language model (GPT-2) is used. 
In our work, we aim at developing a DST model that does not require a large amount of extra resource and can operate efficiently. This is because DST is an intermediate task serving another end task (conversation). To be usable in real scenarios, the DST model should be both time and memory efficient.

\section{Method}
\begin{figure*}[t]
\centering
\includegraphics[height=1.8in]{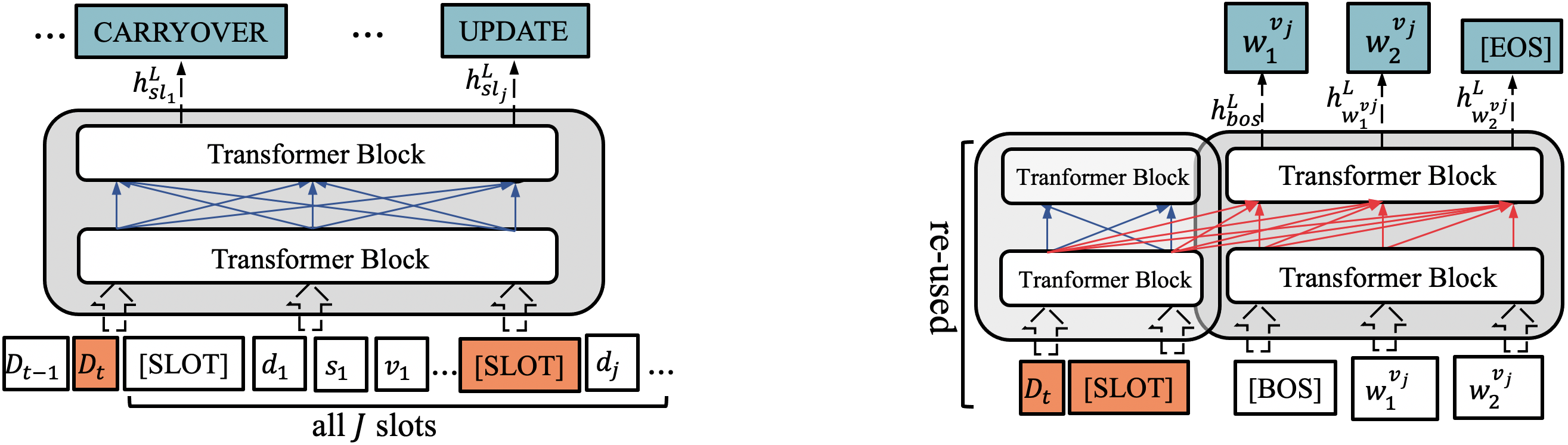}
\caption{(Left) The state operation prediction process, where the model (as the encoder) applies bi-directional self-attention mask. (Right) The value generation process for $j$-th (domain, slot) pair, where the model (as the decoder) applies left-to-right attention and re-uses the hidden states of the encoder in the corresponding decoder layers. The training objective is the sum of the state operation prediction loss and the value generation loss.}
\label{Fig:model}
\end{figure*}

For multi-domain DST, a conversation with $T$ turns can be represented as ${(D_1, S_1), (D_2, S_2), ..., (D_T, S_{T})}$, where $D_t$ is the $t$-th dialogue turn consisting of a system utterance and a user response, $S_{t}$ is the corresponding dialogue state. We define $S_{t}$ as a set of ${(d_j, s_j, v_j)|1 \leq j \leq J}$, where $J$ is the total number of (domain, slot) pairs, i.e. $S_{t}$ records slot values of all (domain, slot) pairs. If no information is given about $(d_j, s_j)$, $v_j$ is \textit{NULL}. 

The goal of DST is to predict $S_t$ given $\{(D_1, S_1), ..., (D_{t-1}, S_{t-1}), (D_t)\}$, i.e. we want to generate the state for the current turn $t$ of dialogue, given the dialogue history and previous dialogue state. Following \citet{kim2019efficient}, we only use $D_{t-1}$, $D_t$, and $S_{t-1}$ to predict $S_{t}$. Figure \ref{Fig:model} gives an overview of our framework. The model is a multi-layer Transformer initialized with BERT, and is used as both the encoder for state operation prediction 
and the decoder for value generation. The differences between the two processes lie in the inputs, self-attention masks and objectives, which will be described in detail.

\subsection{State Operation Prediction}
 
\textbf{Encoder} The input to the encoder is the concatenation of $D_{t-1}$,  $D_t$, and $S_{t-1}$. Each $(d_j, s_j, v_j)$ tuple in $S_{t-1}$ is represented by [SLOT]$\oplus d_j \oplus - \oplus s_j \oplus - \oplus v_j$, where $\oplus$ denotes token concatenation, and [SLOT] and $-$ are separation symbols. Notice that $s_j$ and $v_j$ might consist of several tokens. As illustrated in Figure \ref{Fig:model}, the representations at [SLOT] position $\{\mathbf{x}^L_{sl_j}| 1 \leq j \leq J\}$ are used for state operation prediction. Then, we expect the hidden states at [SLOT] positions are able to aggregate the information from the corresponding $(d, s, v)$ tuples. For example, each $\mathbf{x}^l_{sl_j}$ aggregates information of $(d_j, s_j, v_j)$. 

The input representation, i.e. $\mathbf{X}^0$, is the sum of token embedding, position embedding, and type embedding at each position. We apply type embeddings to introduce a separation between encoder side and decoder side. 
The multi-layer Transformer updates hidden states via: $\mathbf{X}^i={\rm Trans}^i(\mathbf{X}^{i-1}), \quad i \in [1,L]$. Specifically, within a Transformer Block, the multi-head self-attention mechanism is: 
\begin{equation}
\mathbf{C}^l = {\rm Concat}(\mathbf{head}_1, ..., \mathbf{head}_h)
\end{equation}
\begin{equation}
\mathbf{head}_j = {\rm softmax}(\frac{\mathbf{Q}_j\mathbf{K}_j^{T}}{\sqrt{d_k}}+\mathbf{M}^{x})\mathbf{V}_j
\end{equation}
where $\mathbf{Q}_j, \mathbf{K}_j, \mathbf{V}_j \in \mathbb{R}^{n\times d_k}$ are obtained by transforming $\mathbf{X}^{l-1} \in \mathbb{R}^{|x|\times d_h}$ using $\mathbf{W}_{j}^{Q}, \mathbf{W}_j^{K}, \mathbf{W}_j^{V} \in \mathbb{R}^{d_h\times d_k}$ respectively. The self-attention mask matrix $\mathbf{M}^{x} \in \mathbb{R}^{|x|\times |x|}$ (with $\mathbf{M}^{x}_{ij} \in \{0, -\infty \}$) determines whether a position can attend to other positions. Namely, $\mathbf{M}^{x}_{ij}=0$ allows the $i$-th position to attend to $j$-th position and $\mathbf{M}^{x}_{ij}=-\infty$ prevents from it. In the state operation prediction process, $\mathbf{M}^{x}_{ij}=0 \quad \forall i, j$. 

Some hidden states of the encoder will be re-used in the decoder. The outputs of encoder are denoted as $\mathbf{X}^L=[\mathbf{x}^L_{cls}, \mathbf{x}^L_1, ...,\mathbf{x}^L_{sl_1}, ..., \mathbf{x}^L_{sl_J}, ...]$, which will be used for operation prediction.

\textbf{Objective} Following \citet{gao2019dialog} and \citet{kim2019efficient}, we use four discrete state operations: CARRYOVER, DELETE, DONTCARE, and UPDATE. Based on the encoder outputs $\{\mathbf{x}^L_{sl_j}| 1 \leq j \leq J\}$, a MLP layer performs operation classification for each [SLOT].  Specifically, CARRYOVER means to keep the slot value unchanged; DELETE changes the value to NULL; and DONTCARE changes the value to DONTCARE, which means that the slot neither needs to be tracked nor considered important at this turn \cite{wu2019transferable}. Only when UPDATE is predicted does the decoder generate a new slot value for the (domain, slot) pair. 

\subsection{Slot Value Generation}

\textbf{Decoder} It applies different type embeddings to represent its input and left-to-right self-attention mask for generation to avoid information leak. The decoder re-uses \footnote{Re-using means the hidden states do not need to be calculated again in the decoder. In inference, the input of the decoder is only [BOS] to denote the beginning of the string.} the hidden states of encoder in the multi-head self-attention mechanism to construct a flat encoder-decoder structure making parameter updating in the encoder more effective: 
\begin{equation}
\mathbf{Q}_j = \mathbf{Y}^{l-1} \mathbf{W}_{j}^{Q}
\end{equation}

\begin{equation}
\mathbf{\hat{K}}_j = {\rm concat}([\mathbf{\hat{X}}^{l-1}, \mathbf{Y}^{l-1}])\mathbf{W}_{j}^{K}
\end{equation}

\begin{equation}
\mathbf{\hat{V}}_j  = {\rm concat}([\mathbf{\hat{X}}^{l-1}, \mathbf{Y}^{l-1}]) \mathbf{W}_{j}^{V}
\end{equation}

\begin{equation}
\mathbf{head}_j = {\rm softmax}(\frac{\mathbf{Q}_j\mathbf{\hat{K}}_j^{T}}{\sqrt{d_k}}+\mathbf{M}^{y})\mathbf{\hat{V}}_j
\end{equation}
where $\mathbf{Q}_j \in \mathbb{R}^{|y|\times d_k}$, and $\mathbf{\hat{K}}_j, \mathbf{\hat{V}}_j \in \mathbb{R}^{(|\hat{x}|+|y|) \times d_k}$. 
$|\hat{x}|$ is the length of $\mathbf{\hat{X}}$ that is the re-used encoder hidden states . In the decoder, the self-attention mask matrix is $\mathbf{M}^{y} \in \mathbb{R}^{y \times (|\hat{x}|+|y|)}$ and we set $\mathbf{M}^{y}_{ij}=0$ if $j \leq i$. 

Note that the re-used hidden states of the encoder have already encoded the entire input to some extent because of the bi-directional attention applied. We will show in the experiments that re-using only the current turn of dialogue $D_t$ and $j$-th [SLOT], i.e. $\mathbf{x}^l_{sl_j}, l \in \{1, L\}$, (if updating value for the $j$-th slot) achieves the best performance.

\textbf{Objective} The objective of the value generation process is the auto-regressive loss of generated slot values compared to the ground-truth slot values as in the previous works. We use teacher forcing all the time. 
The final training objective is the sum of the state operation prediction loss and the value generation loss.

\section{Experiments}

\begin{table*}[t]
\centering
\small
\begin{tabular}{l|lc|cc}
\hline 
\hline
 & \textbf{Model} & \textbf{BERT used} & \textbf{MultiWOZ 2.0} & \textbf{MultiWOZ 2.1}\\
\hline 
 & HJST \cite{eric2019multiwoz} &  & 38.40 & 35.55 \\
& FJST \cite{eric2019multiwoz} & & 40.20 & 38.00 \\
predefined& SUMBT \cite{lee2019sumbt}  & $\surd$ & 42.40 & - \\
ontology & HyST \cite{goel2019hyst} & & 42.33 & 38.10 \\
 & DS-DST \cite{zhang2019find} & $\surd$ & - & 51.21 \\
& DST-Picklist \cite{zhang2019find} & $\surd$ & - &  53.30 \\
& DSTQA \cite{zhou2019multi} & & 51.44 &  51.17 \\
& SST \cite{chen2020schema} & & 51.17 & 55.23 \\
& CHAN-DST \cite{shan2020contextual} & $\surd$ & \textbf{52.68} & \textbf{58.55} \\
\hline
& DST-Reader \cite{gao2019jointly} &  & 39.41 & 36.40 \\
& DST-Span \cite{zhang2019find} & $\surd$ & - & 40.39 \\
& TRADE \cite{wu2019transferable} & & 48.60 & 45.60 \\
open- & COMER \cite{ren2019scalable} & $\surd$ & 48.79 & - \\
vocabulary & NADST \cite{le2020non} & &  50.52 & 49.04 \\
& SAS \cite{hu2020sas} & & 51.03 &  - \\ 
& SOM-DST \cite{kim2019efficient} & $\surd$ & 51.72 &  53.01 \\
& CSFN-DST \cite{zhu2020efficient} & $\surd$ & 52.23 &  53.19 \\
& Craph-DST \cite{zeng2020multidomain} & $\surd$ & 52.78 &  53.85 \\
& Transformer-DST (ours) & $\surd$ & \textbf{54.64} & \textbf{55.35} \\
\hline
\end{tabular}
\caption{\label{tab:joint} Joint goal accuracy (\%) on the test set of MultiWOZ. Results for the baselines are taken from their original papers. 
}
\end{table*}

\subsection{Datasets}
To evaluate the effectiveness of our approach, we use MultiWOZ 2.0 \cite{budzianowski2018multiwoz} and MultiWOZ 2.1 \cite{eric2019multiwoz} in our experiments. These datasets introduce a new DST task -- DST in mixed-domain conversations. For example, a user can start a conversation by asking to book a hotel, then book a taxi, and finally reserve a restaurant. MultiWOZ 2.1 is a corrected version of MultiWOZ 2.0. Following \citet{kim2019efficient}, we use the script provided by \citet{wu2019transferable} to preprocess the datasets. The final test datasets contain 5 domains, 17 slots, 30 (domain, slot) pairs, and more than 4500 different values. Appendix \ref{app: data} gives more statistics of the datasets. 

\subsection{Implementation Details}
Our model is initialized with BERT (base, uncased), and it works as both the encoder and the decoder. We set the learning rate and warmup proportion to 3e-5 and 0.1. We use a batch size of 16. The model is trained on a P100 GPU device for 15 epochs (a half of the iterations of SOM-DST). We use 42 as the random seed. With it, we can reproduce our experimental results. In the inference, we use the previously predicted dialogue state as input instead of the ground-truth, and we use greedy decoding to generate slot values.

\subsection{Baselines}
We compare the performance of our model, called Transformer-DST, with both ontology-based models and open vocabulary-based models. 

\textbf{FJST} \cite{eric2019multiwoz} uses a bi-directional LSTM to encode the dialogue history and a feed-forward network to choose the value of each slot.

\textbf{HJST} \cite{eric2019multiwoz} encodes the dialogue history using an LSTM like FJST but utilizes a hierarchical network.

\textbf{SUMBT} \cite{lee2019sumbt} uses BERT to initialize the encoder. Then, it scores each candidate slot-value pair using a non-parametric distance measure.

\textbf{HyST} \cite{goel2019hyst} utilizes a hierarchical RNN encoder and a hybrid approach to incorporate both ontology-based and open vocabulary-based settings.

\textbf{DS-DST} \cite{zhang2019find} uses two BERT-based encoders and designs a hybrid approach for ontology-based DST and open vocabulary DST. It defines picklist-based slots for classification similarly to SUMBT and span-based slots for span extraction as DST Reader.

\textbf{DST-Picklist} \cite{zhang2019find} uses a similar architecture to DS-DST, but it performs only predefined ontology-based DST by considering all slots as picklist-based slots.

\textbf{DSTQA} \cite{zhou2019multi} formulates DST as a question answering problem -- it generates a question asking for the value of each (domain, slot) pair. It heavily relies on a predefined ontology.

\textbf{SST} \cite{chen2020schema} utilizes a graph attention matching network to fuse utterances and schema graphs, and a recurrent graph attention network to control state updating.

\textbf{CHAN-DST} \cite{shan2020contextual} employs a contextual hierarchical attention network based on BERT and uses an adaptive objective to alleviate the slot imbalance problem by dynamically adjust the weights of slots during training.


\textbf{DST-Reader} \cite{gao2019dialog} formulates the problem of DST as an extractive question answering task -- it uses BERT contextualized word embeddings and extracts slot values from the input by predicting spans.

\textbf{DST-Span} \cite{zhang2019find} applies BERT as the encoder and then uses a question-answering method similar to DST-Reader.

\textbf{TRADE} \cite{wu2019transferable} encodes the dialogue history using a bi-directional GRU and decodes the value for each state using a copy-based GRU decoder.

\textbf{NADST} \cite{le2020non} uses a transformer-based non-autoregressive decoder to generate the current state.

\textbf{SAS} \cite{hu2020sas} uses slot
attention and slot information sharing
to reduce redundant information’s interference
and improve long dialogue context tracking.

\textbf{COMER} \cite{ren2019scalable} uses BERT-large as the encoder and a hierarchical LSTM decoder.

\textbf{SOM-DST} \cite{kim2019efficient} employs BERT as the encoder and a copy-based RNN decoder upon BERT outputs.

\textbf{CSFN-DST} \cite{zhu2020efficient} introduces the Schema Graph considering relations among domains and slots. Their model is built upon SOM-DST. 

\textbf{Graph-DST} \cite{zeng2020multidomain} introduces the Dialogue State Graph in which domains, slots and values from the previous dialogue state are connected. They instantiate their approach upon SOM-DST for experiments.

\begin{figure*}[t]
\centering
\includegraphics[height=1.2in]{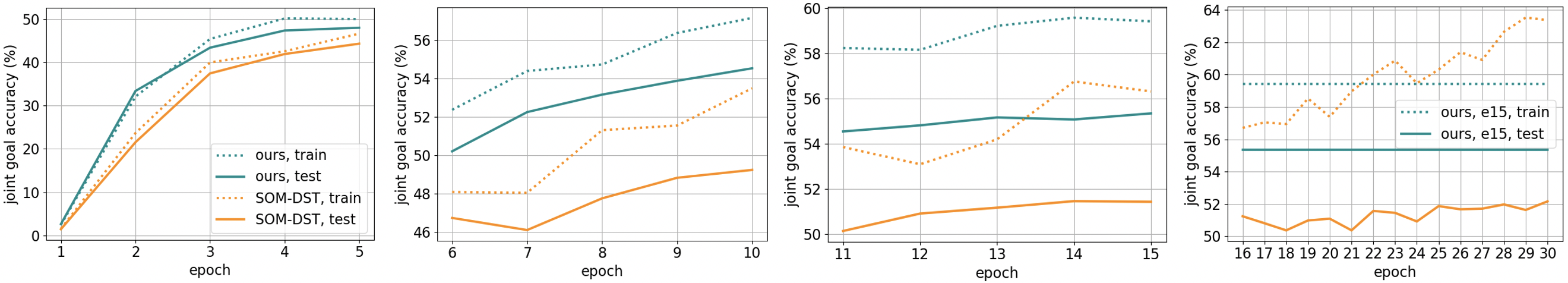}
\caption{The joint goal accuracy of Transformer-DST and SOM-DST on MultiWOZ 2.1.}
\label{Fig:joint}
\end{figure*}

\subsection{Experimental Results}
We report the joint goal accuracy of our model and the baselines on MultiWOZ 2.0 and MultiWOZ 2.1 in Table \ref{tab:joint}. Joint goal accuracy measures whether all slot values predicted at a turn exactly match the ground truth values. The accuracy of baseline models is taken from their original papers.

As shown in the table, our Transformer-DST model achieves the highest joint goal accuracy among open-vocabulary DST: $54.64\%$ on MultiWOZ 2.0 and  $55.35\%$ on MultiWOZ 2.1. Our model even outperforms all ontology-based methods on MultiWOZ 2.0. These latter benefit from the additional prior knowledge, which simplifies DST into a classification/ranking task. 

In Table \ref{tab:domain_jacc}, we show the joint goal accuracy for each domain. The results show that Transformer-DST outperforms previous state-of-the-art framework (SOM-DST) in all domains except in \textit{Taxi}. 
Graph-DST based on SOM-DST introduces the Dialogue State Graph to encode co-occurrence relations between domain-domain, slot-slot and domain-slot. This method outperforms our approach in \textit{Taxi} and \textit{Train} domains. According to the statistics about domains (Appendix \ref{app: data}), we can see that \textit{Taxi} and \textit{Train} frequently co-occur with other domains. Therefore, leveraging extra knowledge about co-occurrence relations is particularly helpful for these domains. Our Transformer-DST does not exploit such knowledge that, however, could be added in the future.

In the following sub-sections, we will examine several questions: 1) how does joint optimization help the model to converge fast? 2) how is the model efficiency? 3) what is the impact of re-using different parts of the model input? 

\begin{table}[t]
\small
\centering
\begin{tabular}{lccccc}
\hline 
\hline
\textbf{Model} & \textbf{Attr.} & \textbf{Hotel} & \textbf{Rest.} & \textbf{Taxi} & \textbf{Train} \\
\hline 
SOM-DST & 69.83 & 49.53 & 65.72 & \textbf{59.96} & 70.36 \\
Graph-DST & 68.06 & 51.16 & 64.43 & 57.32 & \textbf{73.82} \\
Ours & \textbf{71.11} & \textbf{52.01} & \textbf{69.54} & 55.92 & 72.40 \\
\hline
\end{tabular}
\caption{\label{tab:domain_jacc} Domain-specific joint goal accuracy on MultiWOZ 2.1.}
\end{table}

\subsection{Joint Optimization Effectiveness}
\label{sec:joint_optim}

Figure \ref{Fig:joint} shows the joint goal accuracy on training set (5k samples) and test set of each epoch. We train Transformer-DST for 15 epochs and SOM-DST for 30 epochs as suggested in the original paper. 
We can observe that our model performance increases faster than SOM-DST at the beginning, and from 5th to 15th epoch the increase rate of the two frameworks are close. At about 15th epoch, our performance generally stops increasing on both training set and test set.

Our model achieves 54\% joint goal accuracy on the test set at 9th epoch, which already outperforms SOM-DST after 30 epochs. 
In contrast, SOM-DST does not outperform our model (15th epoch) on the training set until 22th epoch. On the test set, SOM-DST performance increases very slowly and is consistently worse than ours. These results suggest that SOM-DST at 30th epoch suffers more from the over-fitting problem than our model. We also observe that its performance fluctuates at the end. 

We also observe that both the training and test curves of our framework are smoother than SOM-DST. 
The same observation is also made 
on MultiWOZ 2.0. This indicates that our training process is more stable and robust.

\subsection{Inference Efficiency Analysis}
\label{sec:effi}

\begin{table}[t]
\centering
\begin{tabular}{lll}
\hline 
\hline
\textbf{Model} & \textbf{Accuracy} & \textbf{Latency} \\
\hline 
TRADE & 45.60 & 450ms \\
NADST & 49.04 &  35ms \\
SOM-DST & 53.01 & 50ms \\
Transformer-DST (Ours) & 55.35 & 210ms \\
\hline
\end{tabular}
\caption{\label{tab:effi} Average inference time per dialogue turn on MultiWOZ 2.1 test set. }
\end{table}

As we have shown that our approach needs much fewer training iterations to achieve state-of-the-art performance, we further analyze its time efficiency at inference/test time. We show in Table \ref{tab:effi} the latency of our method and some typical models measured on P100 GPU with a batch size of 1. Since our approach first predicts state operation, it is about 1 time faster than TRADE that generates the values of all the (domain, slot) pairs at every turn of dialogue. However, Transformer-DST utilizes a multi-layer Transformer (12 layers) for decoding, which makes it 3 times slower than SOM-DST. Overall, when latency is a critical factor in an application, it may be better to use SOM-DST or even NADST (using non-autoregressive decoder). In other cases or having a fast GPU device, the gain in accuracy of Transformer-DST is worth the higher cost in time. More comparison on Inference Time Complexity (ITC) \cite{ren2019scalable} of Transformer-DST and baseline models is provided in Appendix \ref{app:itc}.

\subsection{Resource Required}
\label{sec:resource}

\begin{table}[t]
\centering
\small
\begin{tabular}{llll}
\hline 
\hline
\textbf{Model} & \textbf{Input Len}  & \textbf{Params} & \textbf{Resource} \\
\hline 
Tripy  & 512  & BERT+2xMem & $>$2 \\
SimpleTOD  & 1024 & GPT-2 & $\sim$50  \\
Ours & 256 & BERT & 1 \\
\hline
\end{tabular}
\caption{\label{tab:sota} Efficiency analysis of state-of-the-art approaches via comparing resource usage. BERT (base, uncased) we used has 110M parameters, while GPT-2 has 1.5B parameters.}
\end{table}

The goal of our study is to improve DST without incurring much increase in resources. In parallel, several recent studies have explored using much larger resources for DST. Namely, Tripy \cite{heck2020trippy} and SimpleTOD \cite{hosseini2020simple} also achieve high performance on MultiWoz 2.1 with joint goal accuracy of 55.30\% and 55.76\% respectively. However, these high performances are obtained at the cost of much higher resource requirements. This is why we have not discussed them in previous subsections. Here, we compare our model with these two models in terms of cost-effectiveness.

\paragraph{Extra Knowledge} Tripy uses auxiliary features. Without these extra features, Tripy only obtains 52.58\% joint goal accuracy. SimpleTOD, a multi-task learning approach, also uses extra knowledge, i.e. supervision information from two other tasks closely related to DST. While our approach do not utilize any extra knowledge, the same extra knowledge could also be incorporated and would further boost the performance on DST.

\paragraph{Dialogue History} Both Tripy and SimpleTOD utilize much longer dialogue history as model input, which is important for their models to achieve the state-of-the-art performance as reported. Encoding longer dialogue history is time-consuming and takes several times more GPU memories. 
To avoid this and keep the whole process efficient, we utilize the predicted previous dialogue state as a compact representation of the dialogue history. Even with such a noisy input, our model can still achieve the state-of-the-art performance. 

\paragraph{GPU Usage} According to the input length and the amount of model parameters, we estimated the GPU resources (number of GPUs) needed by each approach as listed in Table \ref{tab:sota}. As mentioned, our approach only needs one P100 GPU for training. In contrast, although SimpleTOD is indeed simple, the GPT-2 it exploited is 13.6 times as larger as BERT we used. The large model is critical to this approach, as the authors also reported that the performance drops if a smaller pre-training model is used instead - with DistilGPT-2, still 8.6 times as larger as our BERT, they only obtained 54.54\% in joint goal accuracy.  

\subsection{Re-Use Hidden States of Encoder}

\begin{table}[t]
\centering
\begin{tabular}{ll}
\hline 
\hline
\textbf{Transformer-DST} & \textbf{Joint Accuracy} \\
\hline 
Full re-use & 27.83 \\
$D_{t-1}$+$D_t$+[SLOT] & 53.08 \\
$D_t$+[SLOT] & \textbf{55.35} \\
$[$CLS$]$+[SLOT] & 53.95 \\
$[$SLOT$]$+(d,s) & 53.83 \\
$[$SLOT$]$ & 53.03 \\
$D_t$+[SLOT]+(d,s,v) & 52.67 \\
$D_t$+[SLOT]+(d,s) & 52.40 \\
\hline
\end{tabular}
\caption{\label{tab:reuse} Joint goal accuracy on MultiWOZ 2.1 by re-using different encoder's states in the decoder.}
\end{table}

In our preliminary experiments, we re-use all hidden states of the encoder in the decoder, and the model performance drops sharply comparing to SOM-DST. Since DST is not a genuine generation task such as dialogue response generation that requires to be consistent with the entire dialogue history or machine translation in which every word on the source side needs to be translated, we consider re-using only a fraction of the hidden states. In SOM-DST, the RNN decoder only uses $\mathbf{x}^L_{cls}$ (the final hidden state at the first position) as the summary of the entire model input and $\mathbf{x}^L_{sl_j}$ (the final hidden state at the $j$-th [SLOT] position) as the summary of the $j$-th (domain, slot, value). Inspired by this, we conduct an exhaustive search on which hidden states should be re-used. The results are listed in Table \ref{tab:reuse}. We can see that re-using encoding hidden states of the current dialogue turn $D_t$ and $j$-th [SLOT] achieves the best performance. Re-using more encoding hidden states may introduce additional noises.

\section*{Conclusion}
The existing state-of-the-art generative framework to DST in open-vocabulary setting exploited BERT encoder and copy-based RNN decoder. The encoder predicts state operation, and then the decoder generates new slot values. However, the operation prediction objective affects only the BERT encoder and the value generation objective mainly influences the RNN decoder because of the stacked model structure. 

In this paper, we proposed a purely Transformer-based framework that uses BERT for both the encoder and the decoder. The operation prediction process and the value generation process are jointly optimized. In decoding, we re-use the hidden states of the encoder in the self-attention mechanism of the corresponding decoder layers to construct a flat encoder-decoder structure for effective parameter updating. Our experiments on MultiWOZ datasets show that our model substantially outperforms the existing framework, and it also achieves very competitive performance to the best ontology-based approaches.  

Some previous works in DST has successfully exploited extra knowledge, e.g. Schema Graph or Dialogue State Graph. Such a graph could also be incorporated into our framework to further enhance its performance. We leave it to our future work.

\bibliography{anthology,acl2020}
\bibliographystyle{acl_natbib}

\clearpage
\appendix

\section{Dataset Statistics}
\label{app: data}
MultiWOZ 2.1 is a refined version of MultiWOZ 2.0 in which the annotation errors are corrected. Some statistics of MultiWOZ 2.1 are reported here.  

\begin{table}[h]
\centering
\begin{tabular}{llll}
\hline 
\hline
\multicolumn{3}{c}{Domain Transition} \\
\cline{1-3}
First & Second & Third & Count \\
\hline 
restaurant & \textbf{train} &  - & 87 \\
attraction & \textbf{train} & - & 80 \\
hotel & -  & - & 71 \\
\textbf{train} & attraction & - & 71 \\
\textbf{train} & hotel & - & 70 \\
restaurant &  - & - & 64 \\
\textbf{train} & restaurant & - &  62 \\
hotel & \textbf{train} &  - &  57 \\
\textbf{taxi} & -  & - &  51 \\
attraction & restaurant &  -  &  38 \\
restaurant &  attraction & \textbf{taxi} & 35 \\
restaurant &  attraction & - & 31 \\
\textbf{train} & - & - & 31 \\
hotel & attraction & - & 27 \\
restaurant & hotel & - &  27 \\
restaurant & hotel & \textbf{taxi} & 26 \\
attraction &  hotel & \textbf{taxi} & 24 \\
attraction & restaurant & \textbf{taxi} & 23 \\
hotel &  restaurant &  - &  22 \\
attraction &  hotel & - & 20 \\
hotel & attraction & \textbf{taxi} & 16  \\
hotel & restaurant & \textbf{taxi} & 10  \\
\hline
\end{tabular}
\caption{\label{tab:domain_transit} Statistics of domain transitions that correspond to more than 10 dialogues in the \textbf{test} set of MultiWOZ 2.1. \textit{Train} domain always co-occurrs with another domain. \textit{Taxi} always co-occurrs with another two domains.}
\end{table}

\begin{table*}[t]
\centering
\begin{tabular}{llccc}
\hline 
\hline
\textbf{Domain} & \textbf{Slots} & \textbf{Train} & \textbf{Valid} & \textbf{Test} \\
\hline 
Attraction & area, name, type & 8,073 & 1,220 & 1,256 \\
Hotel & price range, type, parking, book stay, book day, book people, & 14,793 & 1,781 & 1,756 \\
& area, stars, internet, name &  &  & \\
Restaurant & food, price range, area, name, book time, book day, book people & 15,367 & 1,708 & 1,726 \\
Taxi & leave at, destination, departure, arrive by & 4,618 & 690 & 654 \\
Train & destination, day, departure, arrive by, book people, leave at & 12,133 &  1,972 & 1,976 \\
\hline
\end{tabular}
\caption{\label{tab:stat} Data statistics of MultiWOZ 2.1 including domain and slot types and number of turns in train, valid, and test set. }
\end{table*}

\clearpage

\section{Inference Time Complexity (ITC)}
\label{app:itc}
\begin{table}[h]
\centering
\begin{tabular}{lcc}
\hline 
\hline
\multirow{2}{*}{Model}  & \multicolumn{2}{c}{Inference Time Complexity} \\
\cline{2-3}
 & Best & Worst \\
\hline 
SUMBT & $\Omega(JM)$ & $O(JM)$ \\
DS-DST & $\Omega(J)$ & $O(JM)$ \\
DST-picklist & $\Omega(JM)$ & $O(JM)$  \\
DST Reader & $\Omega(1)$ & $O(J)$ \\
TRADE & $\Omega(J)$ & $O(J)$ \\
COMER &  $\Omega(1)$ & $O(J)$ \\
NADST &  $\Omega(1)$ & $O(1)$ \\
ML-BST &  $\Omega(J)$ & $O(J)$ \\
SOM-DST &  $\Omega(1)$ & $O(J)$ \\
CSFN-DST &  $\Omega(1)$ & $O(J)$ \\
Graph-DST & $\Omega(1)$ & $O(J)$ \\
Transformer-DST (ours) & $\Omega(1)$ & $O(J)$ \\
\hline
\end{tabular}
\caption{\label{tab:itc} Inference Time Complexity (ITC) of our method and baseline models. We report the ITC in both the best case and the worst case for more precise comparison. $J$ indicates the number of slots, and $M$ indicates the number of values of a slot.}
\end{table}

\end{document}